\documentclass{article}

\usepackage{arxiv}

\usepackage[utf8]{inputenc} 
\usepackage[T1]{fontenc}    
\usepackage{hyperref}       
\usepackage{url}            
\usepackage{booktabs}       
\usepackage{amsfonts}       
\usepackage{nicefrac}       
\usepackage{microtype}      
\usepackage{cleveref}       
\usepackage{graphicx}
\usepackage{cite}
\usepackage{doi}
\usepackage{multirow}
\usepackage[dvipsnames,table]{xcolor}
\usepackage{bbding}
\usepackage{authblk}
\usepackage{longtable}
\usepackage{setspace}

\hypersetup{
    colorlinks,
    linkcolor={blue!90!black},
    citecolor={blue!90!black},
    urlcolor={blue!90!black}
}

\title{{\scshape AIvril}: \underline{\smash{AI}}-Driven RTL Generation\\ with \underline{\smash{V}}e\underline{\smash{r}}ification \underline{\smash{I}}n-The-\underline{\smash{L}}oop}

\date{}

\newbox{\orcid}\sbox{\orcid}{\includegraphics[scale=0.06]{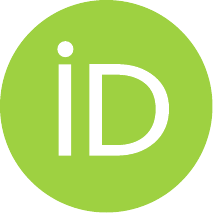}}

\author[1]{%
    Mubashir ul Islam%
}
\author[1]{%
    Humza Sami
}
\author[1,2]{%
    Pierre-Emmanuel Gaillardon
}
\author[1]{%
    Valerio Tenace\thanks{Correspondance: \texttt{valerio@primis.ai}}
}
\affil[1]{PrimisAI, Los Gatos, CA, USA}
\affil[2]{University of Utah, Salt Lake City, UT, USA}

\renewcommand{\headeright}{}
\renewcommand{\undertitle}{}
\renewcommand{\shorttitle}{{\scshape AIvril}: AI-Driven RTL Generation
With Verification In-The-Loop}

\hypersetup{
pdftitle={AIvril: AI-Driven RTL Generation with Verification In-The-Loop},
pdfsubject={cs.AI, cs.MA},
pdfauthor={Mubashir ul Islam, Humza Sami, Pierre-Emmanuel Gaillardon, Valerio Tenace},
pdfkeywords={Large Language Models, RTL Generation, Verification, Multi-Agent Systems, Generative AI, Electronic Design Automation},
}

\begin{document}
\maketitle

\setcounter{footnote}{0}

\begin{spacing}{1}
\begin{abstract}
    {\em Large Language Models} (LLMs) are computational models capable of performing complex natural language processing tasks. Leveraging these capabilities, LLMs hold the potential to transform the entire hardware design stack, with predictions suggesting that front-end and back-end tasks could be fully automated in the near future. Currently, LLMs show great promise in streamlining {\em Register Transfer Level} (RTL) generation, enhancing efficiency, and accelerating innovation. However, their probabilistic nature makes them prone to inaccuracies\textemdash a significant drawback in RTL design, where reliability and precision are essential.
    
    To address these challenges, this paper introduces {\scshape AIvril}, an advanced framework designed to enhance the accuracy and reliability of RTL-aware LLMs. {\scshape AIvril} employs a multi-agent, LLM-agnostic system for automatic syntax correction and functional verification, significantly reducing—and in many cases, completely eliminating—instances of erroneous code generation. Experimental results conducted on the VerilogEval-Human dataset show that our framework improves code quality by nearly 2$\times$ when compared to previous works, while achieving an 88.46\% success rate in meeting verification objectives. This represents a critical step toward automating and optimizing hardware design workflows, offering a more dependable methodology for AI-driven RTL design.
\end{abstract}


\section{Introduction}
In the swiftly advancing domain of {\em Artificial Intelligence} (AI), {\em Large Language Models} (LLMs) have risen as revolutionary tools, possessing the power to transform a multitude of sectors significantly. The specialized sphere of hardware design is no exception as these models have become increasingly influential, particularly due to their proficiency in automating the generation of {\em Register Transfer Level} (RTL) code which could lead to a significant stride towards the automation and optimization of hardware design workflows~\cite{thakur2023verigen,liu2023chipnemo}.

Despite their capabilities, LLMs are not without limitations. Being probabilistic by nature, they are prone to encountering syntax and logical discrepancies, thus mirroring the challenges ubiquitous in traditional programming settings. This issue, however, highlights a critical gap: although LLMs can generate code swiftly and efficiently, the accuracy and functionality of the produced code necessitate thorough verification and refinement. As a byproduct, the benefit of streamlined code generation might be offset by a potentially unsustainable increase in the burden associated with the verification process.

Although still in its infancy, the field of {\em Generative AI} (GenAI) for RTL design has already captured significant interest. Early research focused primarily on the pure code generation capabilities of LLMs~\cite{chang2023chipgpt}. More recent studies have begun to incorporate specific safeguards designed to detect and automatically correct syntax errors~\cite{tsai2023rtlfixer,chang2024data,liu2023chipnemo}, significantly reducing inaccuracies. Yet, as of now, no existing works have fully integrated robust verification mechanisms.

In this paper, we introduce {\scshape AIvril}, a framework that employs a multi-agent approach, integrating automatic syntax correction and a functional verification phase as downstream tasks for RTL-aware language models. This framework leverages the generative capabilities of LLMs to establish a compact yet effective network of intelligent agents. These agents work collaboratively to refine and debug generated code, drawing on feedback from Electronic Design Automation (EDA) tools. Crucially, {\scshape AIvril} interweaves verification with the generative process, thus enhancing the reliability and functionality of the produced RTL code to meet the stringent demands of hardware design. Designed to be tool- and LLM-agnostic, this contribution represents the first significant attempt at creating a {\em GenEDA} framework. As part of the RapidGPT\footnote{\url{https://primis.ai/}} technology stack, {\scshape AIvril} also marks a significant advancement in transparency and usability, offering valuable verification feedback to users\textemdash unlike most existing solutions that operate as ``black boxes''. Experimental results on the VerilogEval-Human dataset highlight {\scshape AIvril}’s effectiveness, demonstrating 1.32$\times$ and 2$\times$ improvements in code quality compared to CodeV~\cite{zhao2024empowering} and RTLFixer~\cite{tsai2023rtlfixer}, respectively. Additionally, it achieves verification goals with an average success rate of 88.46\%, resulting in more robust and compliant RTL implementations.

The remainder of this paper is structured as follows. Section~\ref{sec:background} provides background information and reviews related work in the field, highlighting recent advancements and open challenges. Section~\ref{sec:methods} details the proposed {\scshape AIvril} framework, including implementation details of its underlying architecture. Section~\ref{sec:results} presents the experimental setup and results, demonstrating the effectiveness of {\scshape AIvril} compared to existing approaches. Finally, Section~\ref{sec:conclusions} concludes the paper with a summary of the findings and discusses potential future directions for enhancing AI-driven RTL generation and verification.

\section{Background \& Related Work}\label{sec:background}
Recent advancements in decision-making processes for multi-agent systems have significantly influenced perspective hardware design workflows through the application of GenAI. This section provides an overview of the integration of verbal reasoning and action planning in autonomous systems and examines the growing role of GenAI in RTL design, highlighting both recent progress and ongoing challenges.

\subsection{Decision-Making in Multi-Agent Systems}
Recent work has been focusing on the integration of verbal reasoning and interactive decision-making within autonomous systems, where LLMs have demonstrated to exhibit advanced capabilities in processing multiple reasoning steps to extract answers from various tasks, e.g., arithmetic, commonsense, and symbolic reasoning, thus showing the LLMs' adeptness in navigating intricate reasoning pathways~\cite{wei2022chain}. However, the fact that models tend to rely on their own internal representations, without any external real-world grounding, proved to be a strong limitative factor for reasoning, resulting in increased inaccuracies or hallucination rates as they progress through the chain-of-thought sequence. Other studies have explored LLM agents while executing tasks within interactive environments and then make language-based predictions to formulate action plans~\cite{huang2022language}. In this context, ReAct~\cite{yao2022react} stands out as a pivotal example of this new paradigm. Through reasoning traces, ReAct aids the model in inducing, monitoring, and updating action plans while managing exceptions. Concurrently, actions serve as a medium for the model to interact with and assimilate additional information from external resources, such as knowledge bases, environmental data, or other agents, enhancing its decision-making abilities and contextual awareness. As detailed in the next section, most recent contributions in GenAI for RTL design significantly converge on this paradigm, directly or indirectly, yielding considerable enhancements in the quality of outcomes.

\subsection{Advancements and Challenges of GenAI for RTL Design}\label{sec:related-work}
Chip-Chat~\cite{blocklove2023chip} marked a pioneering effort in emulating a conventional hardware workflow, from design to tapeout, utilizing ``generalist'' LLMs such as ChatGPT throughout the process. Following the introduction of other advanced, open-source LLMs like Llama, CodeGen, and many others~\cite{zhao2023survey}, the realm of RTL code generation has seen a significant boost. Since generating RTL is akin to producing any other programming language, the primary focus has rapidly shifted on enhancing the reliability of these innovative LLM-driven systems. With solutions ranging from domain-adapted LLMs~\cite{liu2023chipnemo} and data augmentation techniques~\cite{chang2024data}, to approaches that integrate an existing knowledge base~\cite{tsai2023rtlfixer} through the Retrieval-Augmented Generation (RAG) mechanism~\cite{lewis2020retrieval}, the current trend leans heavily towards ReAct-based solutions. However, a common limitation among these works is their lack of proper built-in EDA functional verification mechanisms, which negatively affects perceived user feedback. Additionally, many of these solutions rely on fine-tuned models, restricting their adaptability across different settings. This work aims to enhance the reliability of LLM-generated RTL by integrating verification flows within a multi-agent ReAct-based mechanism, thereby providing a more reliable and versatile design experience.

\section{The {\scshape AIvril} Framework}\label{sec:methods}
\begin{figure*}[!tbh]
    \centering
    \includegraphics[width=.9\textwidth]{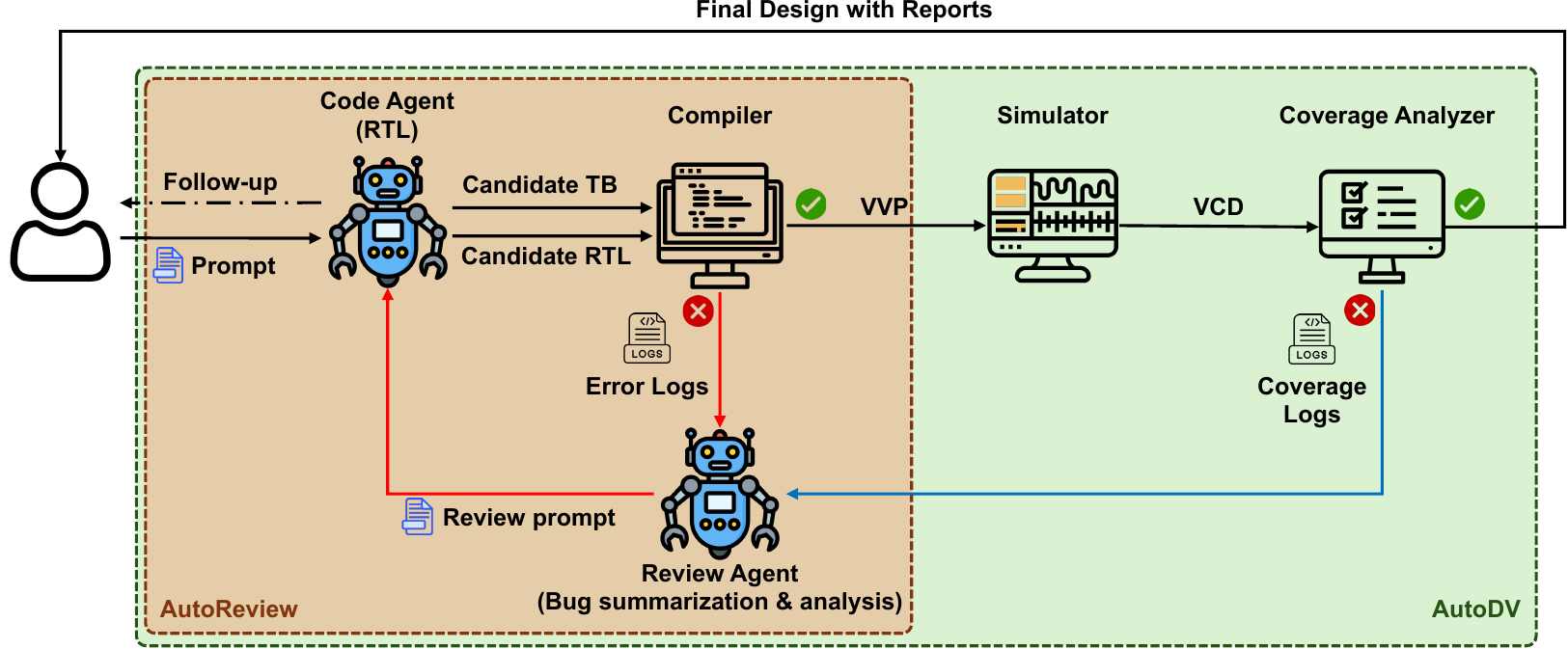}
    \caption{Overall architecture of the proposed {\scshape AIvril} framework.}
    \label{fig:agents}
\end{figure*}

In this section, we introduce the two core components of the {\scshape AIvril} framework, as shown in Figure~\ref{fig:agents}: AutoReview and AutoDV (\underline{{\smash{Auto}}}matic \underline{{\smash{D}}}esign \underline{{\smash{V}}}erification). AutoReview is a design-focused loop that enforces syntax correctness in generated RTL, while AutoDV is a verification loop targeting functional correctness. Both components coordinate automated multi-agent procedures that enhance output quality through systematic iterations and collaborative interactions between agents.

In fact, the effectiveness of {\scshape AIvril} relies on the effective coordination between multiple intelligent agents, each performing specialized tasks. The Code Agent focuses on generating RTL code based on user inputs and corrective feedback, while the Review Agent specializes in error analysis and feedback generation. This multi-agent system operates in a highly coordinated manner, with each agent leveraging its strengths to contribute to the overall goal of producing more accurate RTL code.

Through iterative refinement, the agents collectively reduce the incidence of errors and improve the functional correctness of the code. The integration of functional verification within the generative process distinguishes {\scshape AIvril} from other frameworks, as it provides a more transparent and reliable approach to RTL generation, ultimately leading to higher-quality hardware designs.

\subsection{AutoReview}\label{sec:autoreview}
AutoReview is responsible for enforcing syntax checks and providing automatic corrections for RTL code generated by LLMs. The process begins with the Code Agent, which receives an initial user prompt. These prompts can vary significantly in detail and complexity, and are categorized into three primary scenarios:

\begin{itemize}
    \item {\bf Case I: Detailed Prompts} — In this scenario, the user provides a comprehensive description of the desired design, including specific functionalities and constraints. The Code Agent uses this information to generate candidate RTL implementations and corresponding testbench code. This approach minimizes the need for further interaction, assuming the initial input is complete and precise.

    \item {\bf Case II: Vague Prompts} — When user prompts are broad or lack specifications, the Code Agent engages interactively with the user, asking clarifying questions to gather necessary details. This adaptive querying helps refine the prompt to a level where an actionable RTL design can be generated. In practice, this behavior is realized through {\em ad hoc} system prompts that dynamically adapt the conversation, guiding further inputs to ensure the generated RTL code aligns with the user's implicit requirements.
    
    \item {\bf Case III: Task-Based Prompts} — For prompts that provide a partial or complete RTL description with a request for compilation or verification, the Code Agent utilizes the provided data to generate a testbench and execute the requested tasks. The system is designed to recognize and adapt to varying levels of input detail, ensuring that each task is completed efficiently.    
\end{itemize}

Once a candidate RTL code is produced, it is passed to the syntax check or compilation stage. The Review Agent plays a crucial role here by examining compilation logs for errors. If syntax errors are detected, the Review Agent uses an LLM to analyze the errors, distilling the log data to highlight key issues and generating structured review prompts that provide corrective feedback to the Code Agent. The loop iterates, refining the code with each pass, until all syntax errors are resolved, marking the completion of AutoReview.

\subsection{AutoDV}
AutoDV encapsulates the AutoReview process, therefore concepts introduced in Section~\ref{sec:autoreview} still hold in this scenario. The verification workflow begins with a syntactically correct RTL description, which is subjected to simulation and coverage analysis. The output from these simulations, including coverage reports and any failed assertions, is then analyzed by the Review Agent.

In this phase, the Review Agent identifies discrepancies, such as failed assertions or coverage gaps, and formulates comprehensive review prompts that address both functional and code coverage anomalies. These prompts are fed back to the Code Agent, guiding subsequent iterations of the RTL code. The loop continues until a predefined coverage threshold is achieved, typically set to 90\% or higher to ensure comprehensive verification.

The AutoDV process is tailored to handle various verification challenges, including nuanced aspects of RTL behavior that may not be explicitly stated in the initial design specifications. By iterating between code generation and verification, {\scshape AIvril} ensures that the RTL code not only compiles correctly but it also mitigates functional discrepancies.

\subsection{Integration with EDA Tools and Third-Party LLMs}

{\scshape AIvril} is designed with versatility at its core, being both tool-agnostic and LLM-agnostic. Although this paper focuses on describing specific tools for RTL compilation and coverage analysis, the framework is flexible enough to integrate a broad range of EDA tools, whether open-source or commercial. This allows users to select tools that align with their existing workflows and project requirements without compromising functionality.

In the same vein, {\scshape AIvril} is not dependent on any specific LLM. It can work with various LLMs capable of processing RTL code generation and verification, allowing users to select the model that best fits their requirements.

This dual agnosticism enables {\scshape AIvril} to be seamlessly incorporated into diverse hardware design environments with minimal adaptation, providing a robust and flexible solution that enhances RTL generation and verification across different workflows.

\section{Experimental Results}\label{sec:results}
In this section, we present the experimental evaluation of the proposed {\scshape AIvril} framework. Our objective is to rigorously assess the performance and robustness of our tool across a diverse set of benchmarks, ensuring a fair and comprehensive analysis of its capabilities. To achieve this, we selected key evaluation metrics that highlight the advantages of our approach in realistic and unbiased scenarios. Specifically, we focused on metrics that capture both syntax and functional correctness, as well as success rates in meeting verification objectives, providing a holistic view of {\scshape AIvril}'s effectiveness in design and verification tasks.

\subsection{Methodology}
We employed all the 156 benchmarks from the VerilogEval-Human dataset~\cite{liu2023verilogeval} in our experiments, thus encompassing a wide range of design complexities. For the AutoReview syntax correction loop, we assess performance using the unbiased pass@$k$ estimator (with $k=1$) as described in~\cite{chen2021evaluating}, noting that we differentiate between pass@$k_{syntax}$, which solely indicates the success rate of designs passing all syntax checks, and pass@$k_{functional}$, which reflects the success rate of designs that are not only syntactically correct but also functionally accurate. For AutoDV, we evaluate its robustness by measuring again pass@$k_{functional}$ and its success rate in achieving a predetermined total coverage threshold, set at >90\% for this experiment. This success rate refers to the percentage of benchmarks that met this coverage target, ensuring that at least 90\% of the design's functionality was tested and validated. It is worth noting that in both scenarios, pass@$k_{functional}$ was obtained by executing the testbenches provided by the benchmark suite, ensuring comprehensive validation of the entire approach. For both syntax checks and functional simulation stages, we relied on Icarus Verilog~\cite{williams2002icarus}, whereas Covered~\cite{Chiphackers} was used for coverage analyses.

In our experiments, agents were configured as described in the following:

{\bf \underline{{\smash{Code Agent}}}}---To demonstrate the framework's model-agnostic nature, we resorted to three different LLMs: Claude 3.5 Sonnet~\cite{claude3.5}, GPT-4o~\cite{gpt4o}, and Llama3 70B~\cite{dubey2024llama}. These models were chosen to reflect diverse architectures and capabilities. Importantly, none of these models were fine-tuned specifically for our tasks. On the one hand, this approach eliminates the necessity for any {\em ad hoc} instruction-following fine-tuning, while illustrating, at the same time, how our framework is largely independent of the LLM architecture being used, as long as it possesses adequate RTL knowledge.

{\bf \underline{{\smash{Review Agent}}}}---Similarly, the framework's Review Agent capability was evaluated using the same three LLMs (Claude 3.5 Sonnet, GPT-4o, and Llama3 70B) without any fine-tuning. The review process focused on knowledge distillation from raw output logs, leveraging the models' inherent capabilities to analyze and correct RTL designs effectively.

\begin{figure*}[!tbh]
    \centering
    \includegraphics[width=\textwidth]{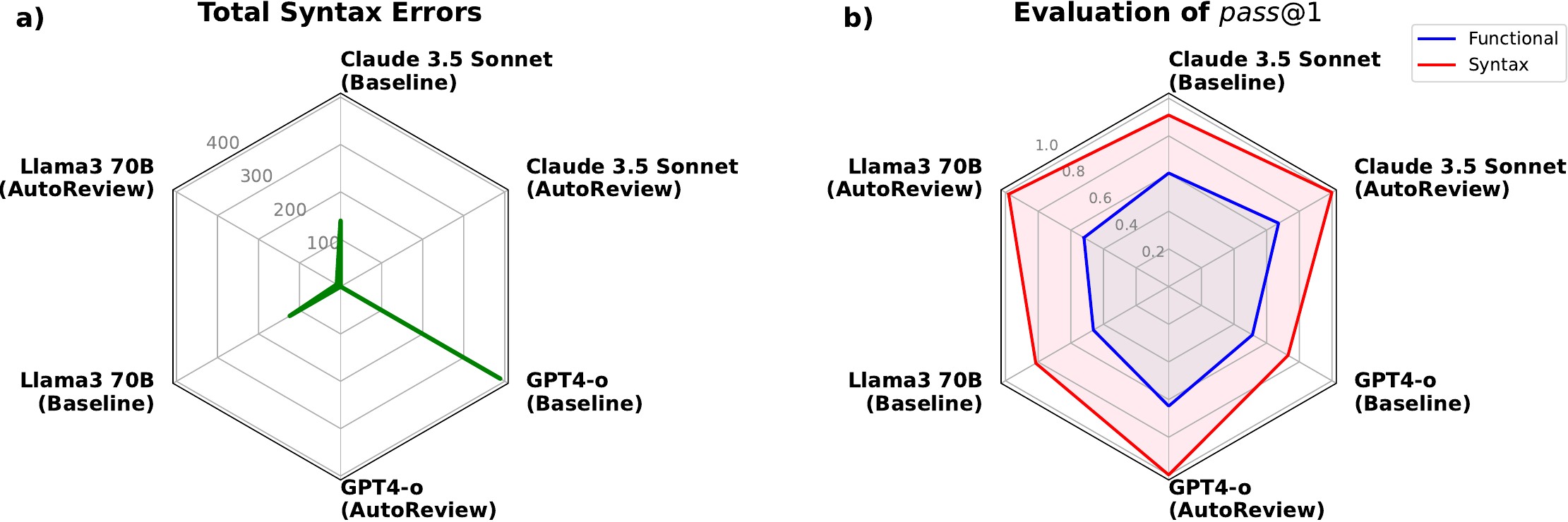}
    \caption{Syntax and functional pass rates for AutoReview. Total number of syntax errors across the benchmark suite (a), and obtained pass@1 scores (b).}
    \label{fig:radar-plots}
\end{figure*}

\subsection{Results \& Discussion}
Figure~\ref{fig:radar-plots} compares the performance of the proposed {\scshape AIvril} framework when using Claude 3.5 Sonnet, GPT4-o, and, Llama3 70B, focusing on the impact of syntax and functional errors reduction through the AutoReview process. Figure~\ref{fig:radar-plots}-a shows the total number of syntax errors for each model under baseline conditions and after applying AutoReview. The results indicate that AutoReview significantly reduces syntax errors across all models. Both GPT4-o and Claude 3.5 Sonnet exhibit complete elimination of syntax errors, dropping from 389 and 139 errors, respectively, to zero after AutoReview. In contrast, Llama3 70B shows a more modest reduction, with syntax errors decreasing from 124 at baseline to a total of 9 after AutoReview.

These results are further highlighted in Figure~\ref{fig:radar-plots}-b, where pass@$k_{syntax}$ (red area) and pass@$k_{functional}$ (blue area) are depicted for each configuration. It is noteworthy how the proposed framework not only enhances syntax correctness but also significantly boosts functional correctness across all benchmarks. The most substantial gains are seen with GTP4-o, where pass@$k_{functional}$ with AutoReview improves over the baseline by 21.2\%. Similarly, Claude 3.5 Sonnet and Llama3 70B exhibit improvements of approximately 11\% and 12\%, respectively.

\begin{figure*}[!tbh]
    \centering
    \includegraphics[width=.75\textwidth]{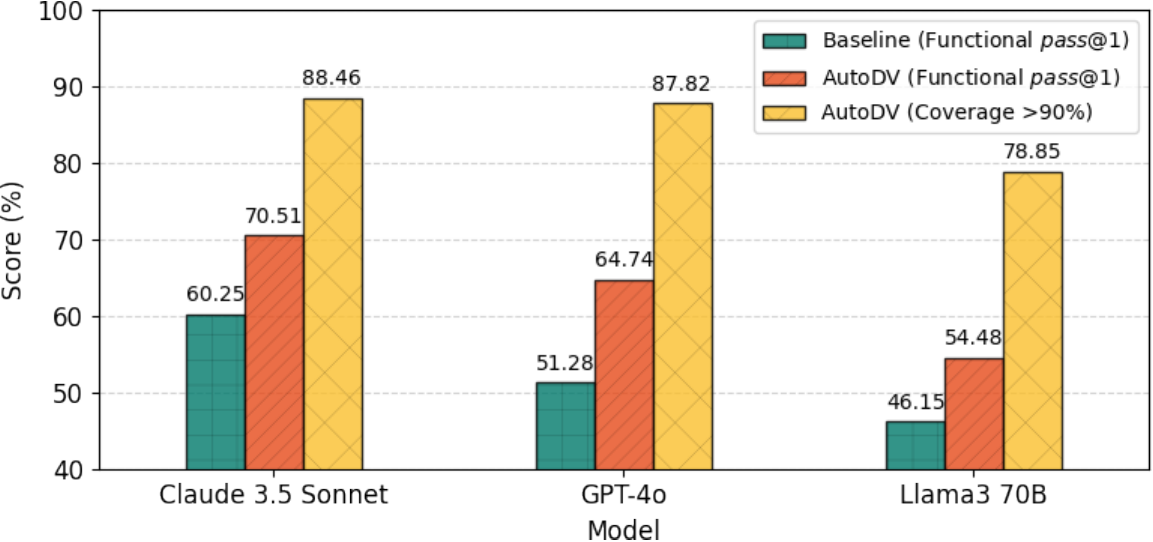}
    \caption{Functional pass rates and verification success rate for AutoDV.}
    \label{fig:bar-plots}
\end{figure*}

Figure~\ref{fig:bar-plots} illustrates the performance of the AutoDV loop. In particular, for each evaluated LLM, we present the pass@$k_{functional}$ for the baseline (green bar), the pass@$k_{functional}$ after applying AutoDV (orange bar), and the success rate in meeting the total coverage target percentage (yellow bar). The plot reveals that AutoDV further enhances the performance of the proposed {\scshape AIvril} framework, resulting in a 23.2\% increase in pass@$k_{functional}$ compared to the baseline for GPT4-o. Consistent with previous findings, Claude 3.5 Sonnet and Llama3 70B show more modest gains of 16\% and 16.5\%, respectively. In terms of meeting the predefined coverage threshold, LLama3 70B recorded the lowest success rate at 78.85\%, while the other two models exceeded a success rate of 87\%.

\subsection{Comparison with State-of-the-Art Approaches}\label{sec}
As highlighted in Section~\ref{sec:related-work}, recent frameworks and fine-tuned LLMs have been proposed to improve RTL code quality in the context of GenAI solutions. RTLFixer~\cite{tsai2023rtlfixer} introduced a ReAct and RAG-based approach that leverages error logs to iteratively correct errors, with a claimed pass@$1_{functional}$ of 36.8\% in the best case. CodeV~\cite{zhao2024empowering} is a collection of open-source instruction-tuned Verilog-driven LLMs. This solution employs fine-tuned versions of CodeLlama 7B Instruct, Deepseek Coder 6.7B Instruct, and CodeQwen1.5 7B Chat on a custom instruction-tuned dataset, achieving a top pass@$1_{functional}$ of 53.2\% on the VerilogEval Human dataset. VerilogCoder~\cite{ho2024verilogcoder} explicitly targets both syntactical and functional errors in RTL code through a multi-agent AI system that autonomously generates and corrects Verilog code. It uses collaborative Verilog tools such as syntax checkers, simulators, and waveform tracers, with the authors reporting a 94.2\% rate of syntactically and functionally correct code on the VerilogEval benchmark dataset. A significant limitation, however, is that VerilogCoder relies on testbenches from the benchmark suite to guide its error correction loop. Although effective in controlled environments, this approach is less practical in real-world scenarios where such gold-standard testbenches are unavailable.

In contrast, the {\scshape AIvril} framework offers a more robust and practical solution by autonomously generating and verifying both RTL code and testbenches, without the need for pre-existing gold-standard blueprints. The integrated AutoReview and AutoDV loops not only significantly reduce syntax errors but also enhance functional correctness across diverse benchmarks. Our findings demonstrate that {\scshape AIvril} consistently outperforms current state-of-the-art methods, i.e., 2$\times$ better than RTLFixer and 1.32$\times$ better than CodeV, thereby improving the robustness and practical applicability of GenAI methodologies for RTL code generation and validation.

\section{Conclusions}\label{sec:conclusions}
The results presented in this paper demonstrate the effectiveness of the proposed {\scshape AIvril} framework in enhancing the accuracy and reliability of GenAI for RTL. By integrating automated review and verification loops—AutoReview and AutoDV—the framework significantly reduces syntax errors and improves functional correctness across various benchmarks. The observed improvements, particularly with GPT4-o, underscore the potential of {\scshape AIvril} to address common pitfalls in LLM-generated code, advancing the state-of-the-art in automated hardware design.

The complete elimination of syntax errors in GPT4-o and Claude 3.5 Sonnet, coupled with notable gains in pass@$k_{functional}$, highlights the robustness of the AutoReview process. Similarly, the AutoDV loop further refines the verification process, achieving significant performance boosts and high success rates in meeting coverage targets. These results validate the framework’s capability to seamlessly integrate syntax correction and verification, enhancing the overall design quality without the need for extensive fine-tuning of the adopted LLMs.

Furthermore, the orthogonality of {\scshape AIvril} to different models and environments, as demonstrated by the diverse performance gains observed with Claude 3.5 Sonnet and Llama3 70B, positions the framework as a versatile and effective GenEDA tool. Future work will focus on further expanding the framework’s verification capabilities and exploring its integration with other LLMs and EDA tools to foster more comprehensive and efficient hardware design consolidation.

In summary, {\scshape AIvril} represents a significant advancement in the field of GenAI for RTL design, offering a reliable, transparent, and efficient approach to overcoming the inherent challenges of LLM-based design automation. Its multi-agent system provides a new benchmark for integrating generative AI with verification mechanisms, paving the way for future innovations in automated hardware design.

\newpage

\bibliographystyle{IEEEtran}
\bibliography{refs}
\end{spacing}

\end{document}